\pdfoutput=1
\documentclass[11pt]{article}

\usepackage[final]{acl}

\usepackage{times}
\usepackage{latexsym}
\usepackage{authblk}
\usepackage{hyperref}
\usepackage{CJKutf8}
\usepackage{times}
\usepackage{graphicx}
\usepackage{amsmath}
\usepackage{booktabs} % For professional quality tables
\usepackage{multirow} % For multirow feature
\usepackage{multicol} % For multicolumn feature
\usepackage{subfig}
\usepackage{CJKutf8}
\usepackage{tabularx}
\usepackage{algorithm}
\usepackage{algpseudocode}
\usepackage{amsmath}

\usepackage[T1]{fontenc}
\usepackage[utf8]{inputenc}

\usepackage{microtype}

\usepackage{inconsolata}

\usepackage{graphicx}
\usepackage{svg}

\title{GuideWeb: A Benchmark for Automatic In-App Guide Generation on Real-World Web UIs}

\author{
\textbf{Chengguang Gan}\textsuperscript{1}\quad
\textbf{Yoshihiro Tsujii}\textsuperscript{1}\quad
\textbf{Yunhao Liang}\textsuperscript{2}
\\
\textbf{Tatsunori Mori}\textsuperscript{4}\quad
\textbf{Shiwen Ni}\textsuperscript{3}\quad
\textbf{Hiroki Itoh}\textsuperscript{1}
\\
\textsuperscript{1}Techtouch, Inc. \\
\textsuperscript{2}University of Chinese Academy of Sciences \\
\textsuperscript{3}Shenzhen Institute of Advanced Technology, Chinese Academy of Sciences \\
\textsuperscript{4}Yokohama National University \\
\small
\ttfamily
\begin{tabular}{c}
\{chengguang.gan, yoshihiro.tsujii, hiroki.itoh\}@techtouch.co.jp \\
liangyunhao22@mails.ucas.ac.cn \quad tmori@ynu.ac.jp \quad sw.ni@siat.ac.cn
\end{tabular}
}

\begin{document}
\begin{CJK}{UTF8}{gbsn}

\maketitle

% \renewcommand{\thefootnote}{\fnsymbol{footnote}}
% \footnotetext[1]{†Corresponding author}

\begin{abstract}

Digital Adoption Platform (DAP) provide web-based overlays that deliver operation guidance and contextual hints to help users navigate complex websites. Although modern DAP tools enable non-experts to author such guidance, maintaining these guides remains labor-intensive because website layouts and functionalities evolve continuously, which requires repeated manual updates and re-annotation. In this work, we introduce \textbf{GuideWeb}, a new benchmark for automatic in-app guide generation on real-world web UIs. GuideWeb formulates the task as producing page-level guidance by selecting \textbf{guide target elements} grounded in the webpage and generating concise guide text aligned with user intent. We also propose a comprehensive evaluation suite that jointly measures the accuracy of guide target element selection and the quality of generated intents and guide texts. Experiments show that our proposed \textbf{GuideWeb Agent} achieves \textbf{30.79\%} accuracy in guide target element prediction, while obtaining BLEU scores of \textbf{44.94} for intent generation and \textbf{21.34} for guide-text generation. Existing baselines perform substantially worse, which highlights that automatic guide generation remains challenging and that further advances are necessary before such systems can be reliably deployed in real-world settings.

\end{abstract}

\section{Introduction}

\begin{figure}[!ht]
  \centering
  \includegraphics[width=\columnwidth]{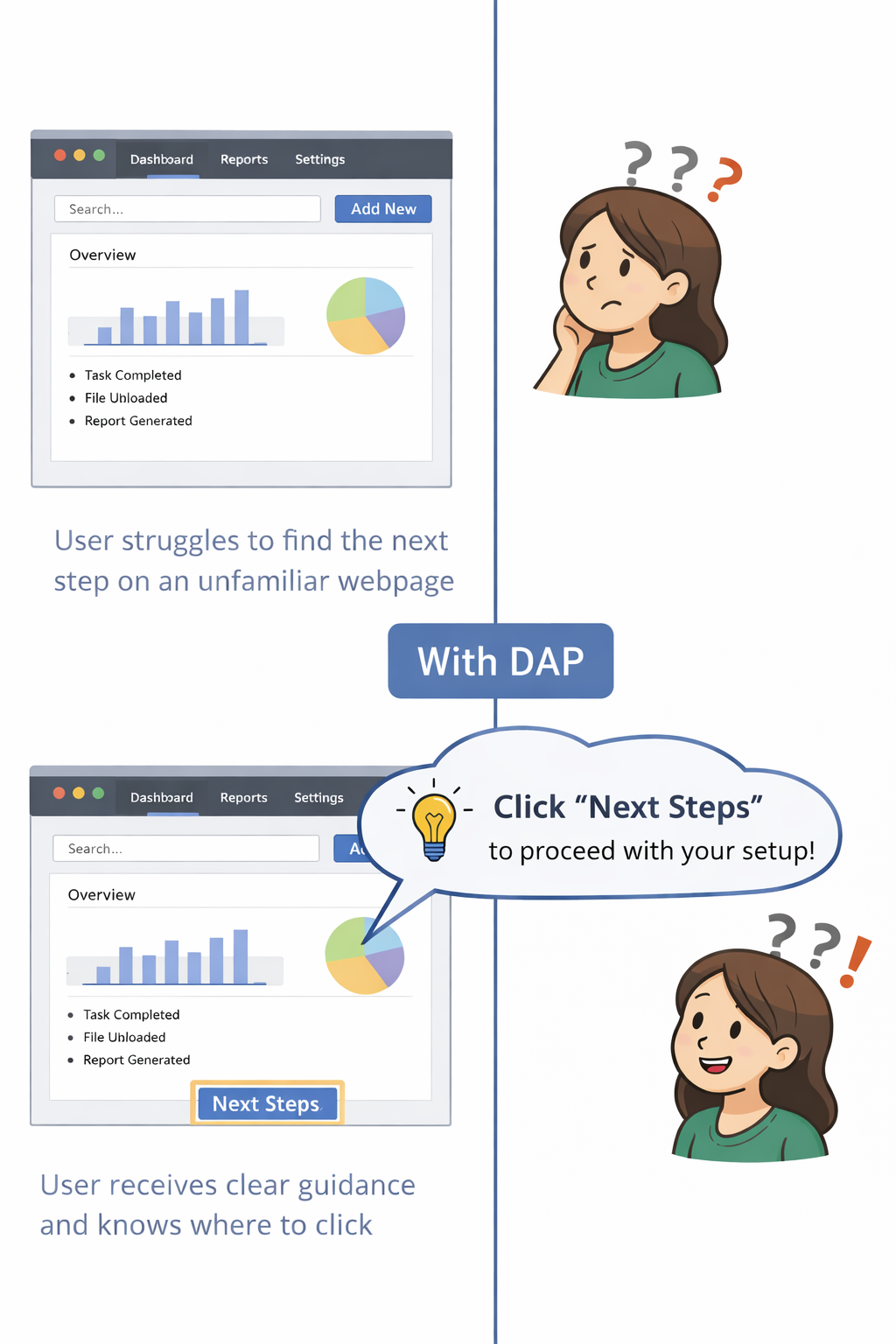}
  \caption{Illustration of a digital adoption platform (DAP) as an in-app overlay for unfamiliar web UIs. 
  \textbf{Top:} without guidance, the user cannot confidently identify the correct guide target element to proceed.
  \textbf{Bottom:} with a DAP overlay, the webpage is augmented with contextual hints and step-by-step instructions, enabling the user to complete the action efficiently.}
  \label{fig:intro_dap}
\end{figure}

\begin{figure*}[!t]
    \centering
    \includegraphics[width=\textwidth]{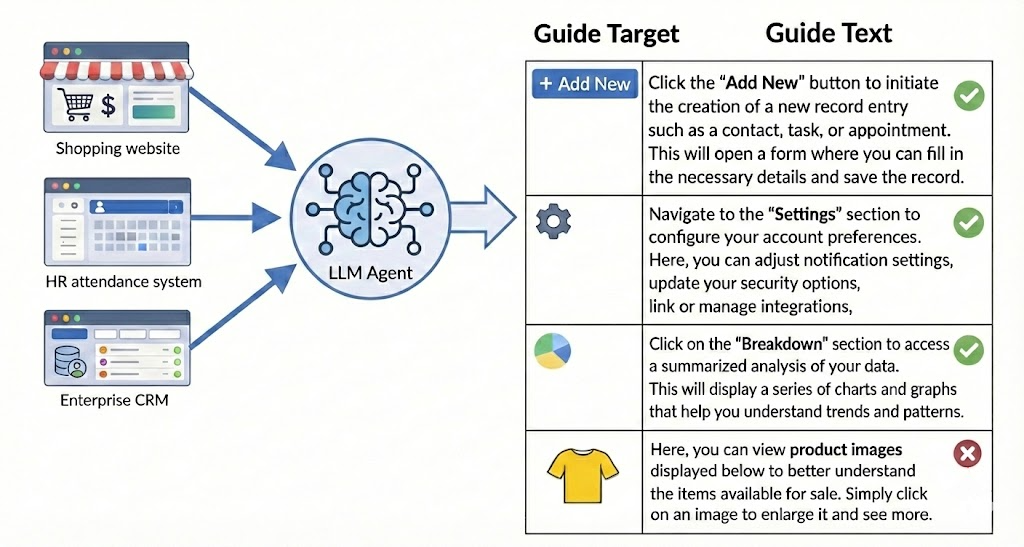}
    \caption{
    Overview of \textsc{GuideWeb}. 
    Given the main page of a real-world website, an LLM-based agent identifies guide targets, namely interactive UI elements whose usage may benefit from guidance, and generates corresponding guide text grounded in visible on-page content. 
    The examples on the right illustrate both correct guides and typical failure cases, where the agent produces low-utility guidance for elements that are already self-explanatory.
    }
    \label{fig:guideweb_overview}
\end{figure*}

In recent years, rapid progress in large language models (LLMs) \citep{openai2025gpt5systemcard,anthropic2025claudeSonnet45SystemCard,comanici2025gemini} has accelerated research on LLM-based agents that plan and act to achieve user goals \citep{huang2024understanding}. Among these, web agents \citep{ning2025survey} have attracted growing attention for their ability to automate interactions with websites and complete complex tasks. While substantial effort has been devoted to benchmarks and methods for action-oriented web agents, an important and widely deployed form of web assistance remains underexplored: in-app guidance for helping users understand and operate unfamiliar web interfaces, as commonly provided by digital adoption platforms (DAPs).

DAPs are typically implemented as browser-based overlays that augment existing web services with contextual hints and step-by-step instructions. As illustrated in Figure~\ref{fig:intro_dap}, users often struggle to identify the correct UI elements to interact with when encountering new or complex webpages, especially in enterprise systems with dense functionality and limited visual cues. This difficulty increases learning cost, leads to inefficient usage, and results in higher customer support burden. By annotating key workflows directly on top of webpages, DAPs enable users to quickly locate and understand essential operations, improving productivity and reducing support cost.

Despite their practical importance, existing DAP solutions rely heavily on manual authoring. Human experts must identify which UI elements require guidance and write corresponding instructional text. This process is costly and brittle, as web layouts and functionalities evolve continuously. Layout changes require re-locating guide targets, and newly introduced features require additional guide creation, while outdated guides can even mislead users. These challenges motivate automatic in-app guide generation with LLM-based web agents.

We address this gap by introducing \textsc{GuideWeb}, a benchmark for automatic web guide annotation and generation grounded in real-world web UIs, as illustrated in Figure~\ref{fig:guideweb_overview}. \textsc{GuideWeb} focuses on main pages of diverse websites and formulates guide generation as a two-stage task: identifying which interactive UI elements should be guided and generating concise guide text aligned with user intent. This formulation reflects a key challenge of in-app guidance: not all interactive elements require explanation, and effective guides must be both selective and informative. To support systematic evaluation, we propose a comprehensive evaluation protocol that measures both guide target selection accuracy and guide text quality.

In addition, we train a lightweight and efficient \textsc{GuideWeb Agent} tailored to this benchmark. Experimental results demonstrate that the trained agent substantially outperforms existing baselines across major evaluation metrics, highlighting both the difficulty of the task and the effectiveness of task-specific modeling for automatic in-app guide generation.

\section{Related Work}

\textbf{Web Agents and Web Interaction Benchmarks.}
Recent advances in large language models have led to rapid progress in web agents that autonomously operate websites to accomplish user goals. Early benchmarks focus on controlled and templated environments, such as \citet{liu2018reinforcement}, which enables reproducible evaluation over synthetic web interfaces. More recent efforts emphasize realism and scale. \citet{yao2022webshop} introduces end-to-end shopping tasks grounded in structured product pages, while \citet{deng2023mind2web} collects diverse real-world trajectories covering a wide range of websites and user intents. \citet{zhou2023webarena} further advances this direction by hosting interactive websites that require multi-step reasoning and form-based interactions across domains. Multimodal extensions such as \citet{koh2024visualwebarena} incorporate visual perception to better model real browsing behavior. Collectively, these benchmarks study how an agent perceives a webpage and executes actions to complete a task. In contrast, our work addresses a different but complementary problem: instead of automating user actions, we focus on generating in-app guidance artifacts that help end users understand how to operate unfamiliar web interfaces.

\textbf{In-App Guidance and UI-Level Assistance.}
Beyond action execution, prior research in human-computer interaction and intelligent assistance has explored how to provide contextual guidance over existing interfaces. Systems such as \citet{zhong2021helpviz} generate visual tutorials aligned with UI states, while approaches like \citet{li2017sugilite} infer task procedures from user demonstrations and UI signals. Other studies investigate interactive walkthroughs, tooltips, and overlay-based explanations to reduce learning cost for complex software. However, these works typically focus on mobile applications, scripted tutorials, or automation by demonstration, and they do not formalize the problem of web-scale guide generation as a benchmarked learning task. Moreover, existing web agent benchmarks do not evaluate whether an agent can identify which UI elements actually require guidance, nor whether the generated explanations are useful to users. GuideWeb fills this gap by introducing a benchmark grounded in real-world web UIs that explicitly decomposes automatic guide generation into guide target identification and guide text generation, together with a comprehensive evaluation protocol for both stages.

\section{GuideWeb: A Benchmark for Automatic In-App Guide Generation on Real-World Web UIs}

We introduce \textsc{GuideWeb}, a benchmark designed to study automatic in-app guide generation on real-world web user interfaces. 
\textsc{GuideWeb} focuses on the main pages of diverse websites and aims to evaluate whether an agent can identify interaction points that require guidance and generate appropriate guide text grounded in visible on-page content.
Unlike prior web agent benchmarks that emphasize long-horizon task completion, \textsc{GuideWeb} targets a complementary but practically important problem: assisting users in understanding and operating unfamiliar web interfaces through lightweight, contextual guidance.

\subsection{Task Definition and Output Schema}

\paragraph{Input Representation.}
Given the main page of a website, we represent the input as the raw HTML source
\begin{equation}
    x \in \mathcal{X},
\end{equation}
from which we construct a DOM tree and extract a set of interactive elements
\begin{equation}
    \mathcal{E}(x) = \{ e_1, e_2, \ldots, e_N \}.
\end{equation}
Each element $e \in \mathcal{E}(x)$ is associated with observable attributes
\begin{equation}
    \phi(e) = (\texttt{tag}(e), \texttt{visible\_text}(e), \texttt{xpath}(e)).
\end{equation}

\paragraph{Output and Task.}
The goal is to generate a structured guide annotation
\begin{equation}
    y = (g, \mathcal{A}),
\end{equation}
where $g \in \{0,1\}$ indicates whether the page requires any guidance, and
$\mathcal{A}$ is a set of element-grounded guide annotations.

Formally, a \textsc{GuideWeb} system is a mapping
\begin{equation}
    f:\mathcal{X} \rightarrow \{0,1\} \times \mathcal{Y}, \quad f(x) = (g, \mathcal{A}).
\end{equation}

\paragraph{Stage 1: Guide Target Identification.}
Conditioned on the page $x$ and its interactive set $\mathcal{E}(x)$, the model selects
a subset of elements that require guidance:
\begin{equation}
    \mathcal{E}^{+} = S(x) \subseteq \mathcal{E}(x).
\end{equation}
Elements in $\mathcal{E}^{+}$ are referred to as \emph{guide targets}.
If the model predicts $g=0$, we define $\mathcal{E}^{+}=\emptyset$ and $\mathcal{A}=\emptyset$.

\paragraph{Stage 2: Element-Grounded Guide Generation.}
For each guide target $e \in \mathcal{E}^{+}$, the model generates a structured annotation
\begin{equation}
    a(e) = (i(e), t(e), s(e), \phi(e)),
\end{equation}

\begin{table*}[!ht]
\centering
\small
\begin{tabular}{p{3.2cm} p{8.5cm}}
\toprule
\textbf{Field} & \textbf{Description} \\
\midrule
\texttt{site} & Unique identifier of the website. \\
\texttt{html\_file} & Raw HTML of the main page (\texttt{page.html}). \\
\texttt{needs\_guides} & Boolean indicating whether the page requires guidance. \\
\texttt{page\_category} & Coarse category of the webpage (e.g., landing, listing). \\
\midrule
\texttt{annotations} & List of guide annotations for selected guide targets. \\
\quad \texttt{intent} & Natural language description of user intent. \\
\quad \texttt{action\_type} & High-level action type (e.g., search, navigation). \\
\quad \texttt{guide\_text} & Generated textual guidance for the target element. \\
\quad \texttt{tag} & HTML tag of the target element. \\
\quad \texttt{visible\_text} & Visible text associated with the element (if any). \\
\quad \texttt{xpath} & XPath locating the target element in the DOM. \\
\bottomrule
\end{tabular}
\caption{Simplified output schema of \textsc{GuideWeb} annotations.}
\label{tab:guideweb_schema}
\end{table*}

where $i(e)$ is a natural-language intent, $t(e)$ is an action type (e.g., search, navigation, selection),
and $s(e)$ is the guide text describing how to use the element. The tuple explicitly carries $\phi(e)$
so that the annotation is grounded in the original page $x$.

The final output set is
\begin{equation}
    \mathcal{A} = \{ a(e) \mid e \in \mathcal{E}^{+} \}.
\end{equation}

This formulation jointly captures \emph{where} guidance is needed (target selection via $S(x)$)
and \emph{what} guidance should be provided (element-conditioned generation of $a(e)$),
while grounding all outputs in the input webpage through $\texttt{xpath}$ and visible content.

\paragraph{Output Schema.}
Each annotated webpage in \textsc{GuideWeb} is stored in a structured JSON format.
The schema consists of page-level fields and a list of guide-level annotations, summarized in Table~\ref{tab:guideweb_schema}.

As summarized in Table \ref{tab:guideweb_schema}, each webpage in \textsc{GuideWeb} is stored in a dedicated subdirectory.
The directory contains the raw HTML file of the webpage main page, as well as a JSON file that stores all annotation-related information.
At the page level, the JSON file records whether the webpage requires guidance and includes coarse-grained category information.
At the guide level, the file contains a collection of annotated guide targets, each corresponding to an interactive element that requires guidance.

For each guide target, we store the element’s visible text, namely the textual content presented to users on the webpage.
We note that some interactive elements are purely icon-based and do not expose visible text; in such cases, the element is referenced solely by its XPath.
The XPath field enables precise localization of the target element within the DOM, which is essential for downstream DAP systems to reliably attach guidance overlays to the correct UI components.

In addition to grounding information, we annotate each guide target with an action type (e.g., search, login), which captures the high-level functional category of the interaction.
The \texttt{tag} field records the raw HTML tag type of each guide target element as it appears in the DOM.
For example, many interactive controls in real-world websites are implemented using anchor tags (\texttt{<a>}), even when they function visually as buttons, tabs, or menu items.
Therefore, \texttt{tag} reflects the structural implementation of an element rather than its semantic action type, and is used primarily for grounding guide annotations to concrete DOM nodes.

Finally, we associate each guide target with a natural-language intent description.
The intent serves two purposes.
First, it encourages the LLM to explicitly reason about why a particular guide is needed and what user goal it supports, thereby improving the accuracy of guide target identification.
Second, the intent enables downstream DAP systems to proactively confirm whether the user shares the same goal, facilitating faster and more targeted assistance during interaction.

\begin{figure}[!h]
    \centering
    \includegraphics[width=\columnwidth]{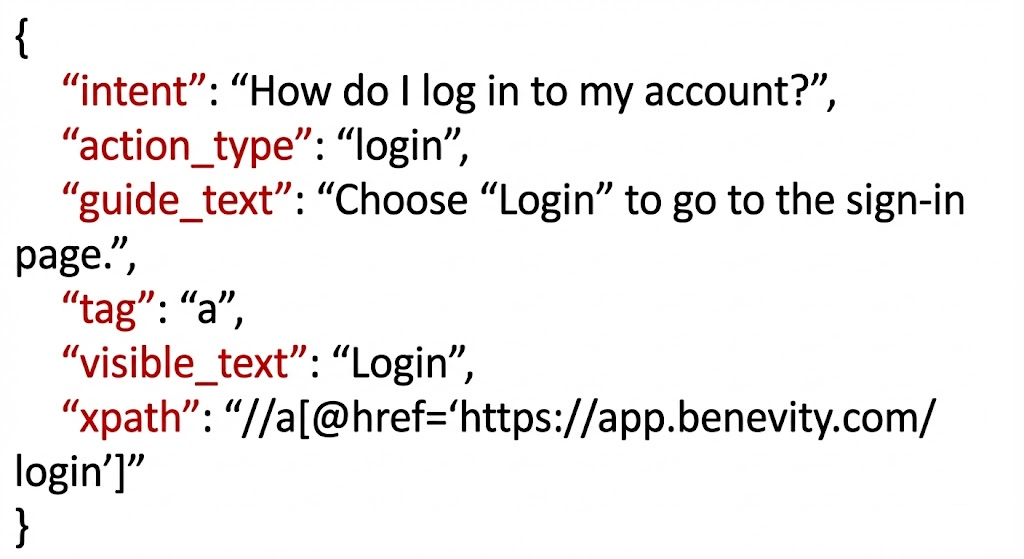}
    \caption{
    A single guide annotation in \textsc{GuideWeb}, containing intent, action type, guide text, and DOM grounding fields.
    }
    \label{fig:guideweb_json_example}
\end{figure}

\begin{figure*}[!ht]
    \centering
    \includegraphics[width=\textwidth]{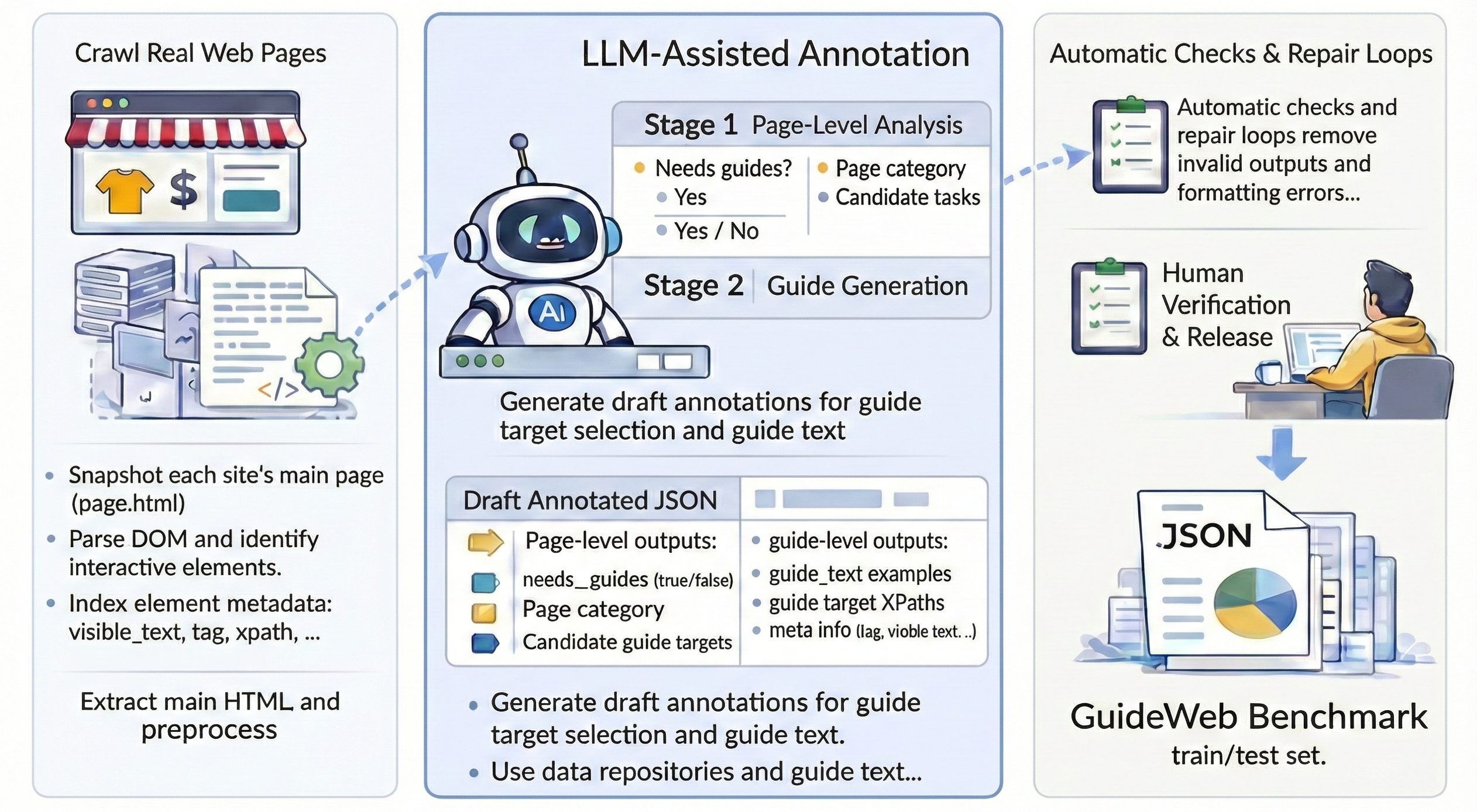}
    \caption{Overview of the \textsc{GuideWeb} construction pipeline with LLM-assisted annotation and human verification.}
    \label{fig:guideweb_construction}
\end{figure*}

As illustrated in Figure \ref{fig:guideweb_json_example}, all of the above information is unified within a single JSON file, allowing models to generate, store, and consume annotations in a consistent and structured format.

\subsection{Dataset Construction Pipeline}

During dataset construction, we adopt a hybrid annotation strategy that combines LLM-assisted labeling with human verification and correction.
The overall pipeline of constructing \textsc{GuideWeb} is illustrated in Figure~\ref{fig:guideweb_construction}.
This design allows us to efficiently scale the annotation process while maintaining high annotation quality.

The websites included in \textsc{GuideWeb} are sampled from the Cisco Umbrella Popularity List,\footnote{\url{https://umbrella-static.s3-us-west-1.amazonaws.com/index.html}}
which ranks domains based on aggregated passive DNS usage observed across Umbrella’s global network.
Unlike browser-centric rankings such as Alexa, which rely primarily on HTTP traffic collected from browser plugins, the Umbrella list reflects Internet-wide activity across diverse protocols and applications.
This property makes it particularly suitable for collecting a broad and realistic set of web services.

We begin with the Top 1M domains provided by the Umbrella list\footnote{\url{http://s3-us-west-1.amazonaws.com/umbrella-static/top-1m.csv.zip}} and randomly sample candidate domains.
For each selected domain, we use an automated browser to crawl and snapshot its main landing page.
We retain only websites whose main pages satisfy minimum structural requirements, such as containing a sufficient number of interactive elements (e.g., clickable controls, input fields, and forms).
This filtering step ensures that each retained page presents meaningful opportunities for in-app guidance generation.
After filtering, we obtain a final set of 1,000 real-world websites spanning diverse categories, including e-commerce platforms, enterprise systems, and general service websites.

For each selected website, the main page serves as the annotation target.
We first store the raw HTML snapshot of the page and parse its DOM structure to identify interactive elements.
Each interactive element is indexed using a unified \texttt{elements index}, which records its raw HTML tag type (\texttt{tag}), visible text content (\texttt{visible\_text}), and a precise XPath (\texttt{xpath}) for reliable localization.
Notably, some interactive elements, such as icon-only buttons, do not expose visible text and are therefore indexed using their structural attributes alone.
This element indexing step forms the foundation for grounding all subsequent guide annotations to concrete DOM nodes.

The processed and indexed HTML is then provided to an LLM for assisted annotation.
At the first stage, the LLM is instructed to perform page-level analysis, including determining whether the page contains interactive elements that merit guidance and assigning a coarse-grained page category.

\begin{table*}[!ht]
\centering
\small
\setlength{\tabcolsep}{4pt}
\begin{tabular}{l r @{\hskip 10pt} l r}
\toprule
\textbf{Statistic} & \textbf{Value} & \textbf{Statistic} & \textbf{Value} \\
\midrule
Total crawled websites & 1{,}000
& Websites removed after verification & 4 \\
Valid annotated websites & 996
& Websites requiring guides (\texttt{needs\_guides=true}) & 980 \\
Websites without guides & 16
& Ratio requiring guides & 98.4\% \\
\midrule
Average guides per page & 3.09
& Pages reaching guide cap (5 guides) & 546 (54.8\%) \\
\bottomrule
\end{tabular}
\caption{Overall statistics of the \textsc{GuideWeb} benchmark (guide-level annotations only).}
\label{tab:guideweb_overall}
\end{table*}

If the page is judged to require guidance, the LLM proceeds to identify a subset of interactive elements that should be annotated as guide targets.
At the second stage, the LLM generates corresponding guide text for each selected target, along with an explicit user intent describing the underlying user goal.
In this formulation, the guide text serves as an answer-like explanation, while the intent captures the implicit question a user may have when interacting with the page.
All annotations are generated in a predefined JSON schema specified in the prompt.
If the model output does not conform to the required format, an automatic regeneration and repair loop is triggered until a valid structured output is obtained.

The LLM-generated annotations are subsequently reviewed and corrected by human annotators with higher education backgrounds.
Each annotation is verified against the original webpage to ensure correctness, clarity, and practical usefulness.
Through this verification process, we remove four pages with severe structural errors or annotation ambiguities.
The final benchmark therefore consists of 996 valid annotated webpages.
We split these samples into training and test sets with a ratio of 7.5:2.5, which are used consistently across all experiments.

\subsection{Dataset Statistics}

Table~\ref{tab:guideweb_overall} reports high-level benchmark statistics. Starting from 1{,}000 crawled domains, we obtain 996 validated main-page snapshots after human verification. The vast majority of pages are deemed guide-worthy (\texttt{needs\_guides=true}), reflecting that real-world homepages typically expose multiple interactive entry points that benefit from lightweight in-app guidance. On average, each page contains 3.09 guide annotations, and more than half of the pages hit the per-page annotation budget of five guides, indicating that even a conservative cap is frequently saturated in practice. This design choice reflects the observation that densely annotating a page with excessive guidance is neither necessary nor desirable, as it may hinder usability and visual clarity; the limit of five guides is therefore introduced in an exploratory setting to prioritize the most valuable, frequently used, and user-critical interactions.

\begin{table}[!t]
\centering
\small
\setlength{\tabcolsep}{6pt}
\begin{tabular}{l r}
\toprule
\textbf{Action type} & \textbf{\# Guides} \\
\midrule
search & 728 \\
navigate & 520 \\
login & 412 \\
contact\_support & 307 \\
other (incl. $<$50 types) & 296 \\
signup & 199 \\
subscribe\_newsletter & 112 \\
start\_trial & 107 \\
checkout & 93 \\
pricing & 86 \\
filter\_sort & 79 \\
download\_install & 75 \\
settings\_profile & 64 \\
\bottomrule
\end{tabular}
\caption{Distribution of guide annotations by action type (guide-level counts). Rare types with fewer than 50 guides are merged into \texttt{other}.}
\label{tab:guideweb_action_types}
\end{table}

Table~\ref{tab:guideweb_action_types} summarizes the distribution of guide annotations by action type. The benchmark is dominated by information-seeking and navigation behaviors (e.g., \texttt{search} and \texttt{navigate}), followed by common account and support workflows (e.g., \texttt{login}, \texttt{signup}, \texttt{contact\_support}). Long-tail categories are aggregated into \texttt{other} to stabilize analysis and avoid over-interpreting sparse types. A more detailed breakdown of page categories and their relationship with guide necessity is provided in Appendix~\ref{app:page_category_analysis}.

\section{GuideWeb Agent}

\begin{figure}[t]
    \centering
    \includegraphics[width=\columnwidth]{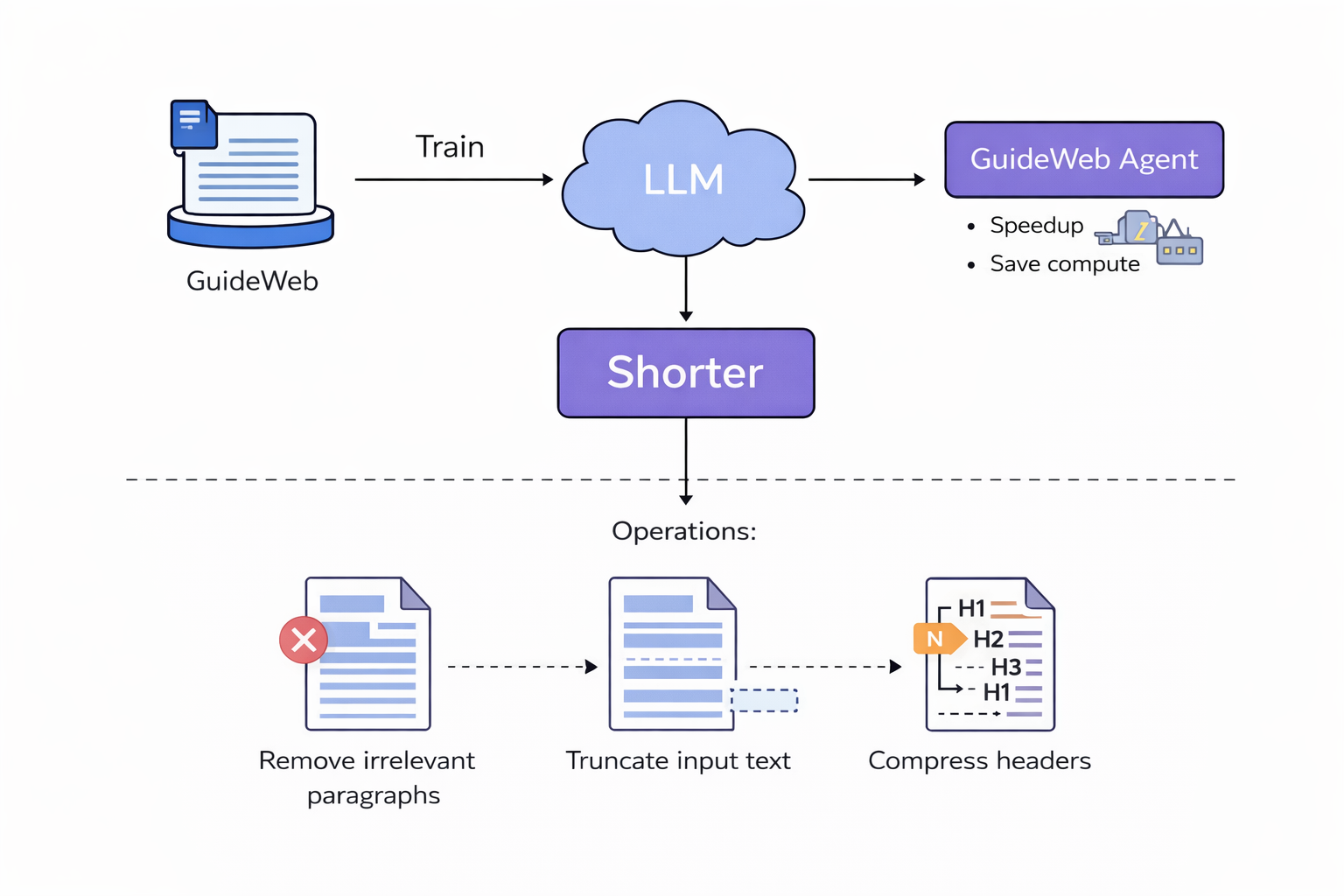}
    \caption{
    Overview of the \textsc{GuideWeb Agent} with the \textit{Shorter} mechanism.
    The agent is trained on the \textsc{GuideWeb} benchmark to perform automatic in-app guide generation.
    During inference, the \textit{Shorter} module reduces input length by removing irrelevant content, truncating long text, and compressing headers, enabling faster inference and reduced computational cost while preserving guide-relevant information.
    }
    \label{fig:guideweb_agent}
\end{figure}

In addition to the benchmark, we train a lightweight \textsc{GuideWeb Agent} on the GuideWeb training set for automatic in-app guide generation.
To address the challenge of long and noisy webpage inputs, we incorporate a \textit{Shorter} mechanism that removes irrelevant HTML content and compresses the remaining structure before inference.
This design substantially reduces input length, leading to faster inference and lower computational cost, while preserving the information necessary for guide target identification and guide text generation.

Empirical results in Section~\ref{sec:results} show that the \textit{Shorter} mechanism does not degrade performance and in some cases improves it.
These findings suggest that, for web guide generation, the information required by the model is largely concentrated in interactive elements, their visible texts, and nearby contextual descriptions.
In contrast, large portions of raw webpage text and auxiliary HTML structures contribute little to the task.
This observation supports the use of compact, interaction-focused representations for efficient and effective in-app guide generation.

\section{Experiments}

\subsection{Evaluation Metrics}
\label{sec:metrics}

We evaluate \textsc{GuideWeb} from three complementary aspects:
(1) \textbf{guide target selection} (which UI elements should be annotated),
(2) \textbf{text generation quality} (intent and guide text), and
(3) \textbf{structured grounding accuracy} (field-level exact matches).
While the underlying measures are standard, we tailor their evaluation targets to the \textsc{GuideWeb} setting.

\paragraph{Guide target selection (P/R/F1).}
For each page, let $G$ be the set of gold guide targets and $\hat{G}$ be the predicted set, where each target corresponds to a unique DOM element.
We report precision, recall, and F1 over set overlap (definitions in Appendix~\ref{app:metrics}).
Intuitively, precision measures how many predicted targets are truly guide-worthy, recall measures how many gold targets are recovered, and F1 summarizes the trade-off.
This metric captures \emph{selective} guide authoring, penalizing redundant or low-utility annotations.

\paragraph{Intent and guide-text generation (BLEU, ROUGE-L).}
For matched guide targets, we compare generated intent strings and guide texts against references using BLEU and ROUGE-L (standard definitions).
These metrics quantify lexical overlap and sequence-level similarity, respectively, and reflect whether the agent produces concise, faithful descriptions aligned with the intended interaction.

\paragraph{Structured field correctness (Exact-match F1).}
We additionally evaluate whether predictions are grounded in the webpage structure by computing exact-match F1 for key fields:
\texttt{action\_type}, \texttt{tag}, \texttt{visible\_text}, and \texttt{xpath}.
Here, a field value is counted as correct only if it exactly matches the reference.
This provides fine-grained diagnostic signals: \texttt{xpath} tests precise localization, whereas \texttt{tag} and \texttt{visible\_text} reflect semantic and surface alignment.

\subsection{Baselines}

We evaluate \textsc{GuideWeb} using a diverse set of representative large language models as baselines, covering both proprietary and open-source systems.
Specifically, we include GPT-5~\citep{singh2025openaigpt5card}, Claude Sonnet 4.5~\citep{anthropic2025claudeSonnet45SystemCard}, and Gemini 2.5 Pro~\citep{comanici2025gemini} as strong closed-source LLM baselines, as well as Qwen3-8B~\citep{yang2025qwen3technicalreport} and LLaMA 3.1-8B~\citep{grattafiori2024llama3herdmodels} as competitive open-source counterparts.

All baseline models are prompted to perform guide target identification and guide text generation directly from webpage content under a unified prompt template, without task-specific fine-tuning.
This setting reflects a realistic zero-shot or instruction-following deployment scenario and allows us to isolate the inherent capability of general-purpose LLMs on the \textsc{GuideWeb} benchmark.

\section{Results}\label{sec:results}

\begin{table*}[t]
\centering
\caption{Main results on GuideWeb. Left: guide target element selection. Right: intent and guide-text generation quality.}
\label{tab:main_results}
\resizebox{\textwidth}{!}{%
\begin{tabular}{lrrrrrrrr}
\toprule
& \multicolumn{4}{c}{\textbf{(a) Guide target element selection}} & \multicolumn{4}{c}{\textbf{(b) Text generation quality}} \\
\cmidrule(lr){2-5}\cmidrule(lr){6-9}
\textbf{Model} & \textbf{P} & \textbf{R} & \textbf{F1} & \textbf{Match/Pred.}
& \textbf{Intent BLEU} & \textbf{Intent ROUGE-L} & \textbf{Guide BLEU} & \textbf{Guide ROUGE-L} \\
\midrule
GPT-5 & 15.29 & 54.69 & 23.90 & 356/2328 & 13.84 & 24.99 & 4.30 & 11.69 \\
Claude-Sonnet-4.5 & 15.99 & \textbf{58.99} & 25.16 & 384/2402 & 15.88 & 25.99 & 1.48 & 8.98 \\
Gemini-2.5-Pro & 14.43 & 54.84 & 22.85 & 357/2474 & 16.53 & 30.54 & 2.53 & 13.43 \\
Qwen3-8B & 9.11 & 12.33 & 10.48 & 72/790 & 3.69 & 11.54 & 1.93 & 11.47 \\
LLaMA3.1-8B & 10.98 & 2.77 & 4.42 & 18/164 & 4.13 & 13.57 & 2.05 & 11.45 \\
\midrule
\textbf{GuideWeb Agent} & \textbf{29.99} & 31.64 & \textbf{30.79} & 206/687
& \textbf{44.94} & \textbf{52.89} & \textbf{21.34} & \textbf{28.44} \\
\bottomrule
\end{tabular}%
}
\end{table*}

As shown in Table\ref{sec:results}, we evaluate all baseline models and the proposed GuideWeb Agent using the comprehensive metric suite introduced in Section~\ref{sec:metrics}. We first focus on guide target element selection. The GuideWeb Agent achieves the highest F1 score among all models, indicating a substantially better balance between precision and recall. Although several baseline models exhibit much higher recall than the trained GuideWeb Agent, this does not imply superior performance. Instead, these models achieve high recall by producing a significantly larger number of predictions, effectively relying on over-generation to increase the chance of matching gold guide targets.

This behavior is clearly reflected in the \emph{Match/Predicted} statistics. Compared to the GuideWeb Agent, baseline models generate approximately three to four times more predicted guide targets, yet they only identify about 1.5× more correct matches. Such a strategy inflates recall at the cost of precision and indicates that these models fail to selectively identify guide-worthy elements. Rather than understanding which interactive elements truly require guidance, they tend to mark many elements indiscriminately, resulting in redundant and low-utility guides.

In contrast, for both user intent generation and guide text generation, the trained GuideWeb Agent consistently outperforms all baselines by a large margin across BLEU and ROUGE-L metrics. This demonstrates that task-specific training enables the agent to better align generated intents and guide texts with the underlying page semantics and user needs. Overall, these results highlight that existing general-purpose models struggle to solve the GuideWeb task effectively, and that dedicated modeling and training are crucial for accurate and practical in-app guide generation.

\section{Ablation of the Shorter Mechanism}

\begin{table}[t]
\centering
\caption{Ablation of the Shorter mechanism.}
\label{tab:ablation_shorter}
\resizebox{\columnwidth}{!}{%
\begin{tabular}{lrrrr}
\toprule
\multicolumn{5}{c}{\textbf{Qwen3-8B}} \\
\midrule
\textbf{Setting} & \textbf{P} & \textbf{R} & \textbf{F1} & \textbf{Match/Pred.} \\
\midrule
w/o Shorter & 9.11 & 12.33 & 10.48 & 72/790 \\
Shorter & \textbf{9.52} & \textbf{29.67} & \textbf{14.42} & 181/1901 \\
\bottomrule
\end{tabular}%
}
\end{table}

Table \ref{tab:ablation_shorter} reports an ablation study on the proposed \emph{Shorter} mechanism using Qwen3-8B as the backbone model. Without Shorter, the model processes raw webpage inputs with substantially longer HTML context, resulting in limited recall and overall F1 performance. After enabling the Shorter mechanism, which selectively preserves interactive elements and nearby informative text, recall improves markedly from 12.33 to 29.67, leading to a consistent F1 gain.

Notably, this improvement is achieved without increasing model capacity or modifying training objectives. Instead, Shorter reduces redundant and low-utility HTML content, allowing the model to focus on semantically relevant signals for guide target selection. These results indicate that effective context reduction is critical for GuideWeb tasks, and that long-form raw HTML is not only inefficient but can actively hinder accurate guide prediction.

\section{Conclusion and Future Work}

We introduce \textsc{GuideWeb}, the first benchmark for automatic in-app guide generation on real-world web UIs. GuideWeb formalizes guide generation as selecting guide-worthy interactive elements and generating concise, user-aligned guide text. We design a comprehensive evaluation suite and propose a lightweight \textsc{GuideWeb Agent} with a context-shortening mechanism that significantly outperforms strong general-purpose baselines.

Our results show that existing models tend to over-generate guide targets to boost recall, while failing to selectively identify meaningful guidance. In contrast, task-specific training and input structuring enable more precise and informative guide generation. Future work includes extending GuideWeb to multi-step workflows and exploring tighter integration with real-world DAP systems.

\section{Limitations}

This work focuses exclusively on main pages of web applications and does not address multi-page or stateful workflows that require long-horizon planning. In addition, although GuideWeb leverages human verification to ensure annotation quality, the benchmark remains limited in scale compared to fully automated web interaction datasets. Finally, our GuideWeb Agent is trained in an offline setting and does not adapt to user feedback or evolving webpage layouts at inference time. We leave interactive learning, broader page coverage, and online adaptation to future work.

% \section*{Acknowledgments}

\bibliography{custom}

\appendix

\section{Page Category Analysis}
\label{app:page_category_analysis}

\begin{figure*}[!ht]
    \centering
    \includegraphics[width=\textwidth]{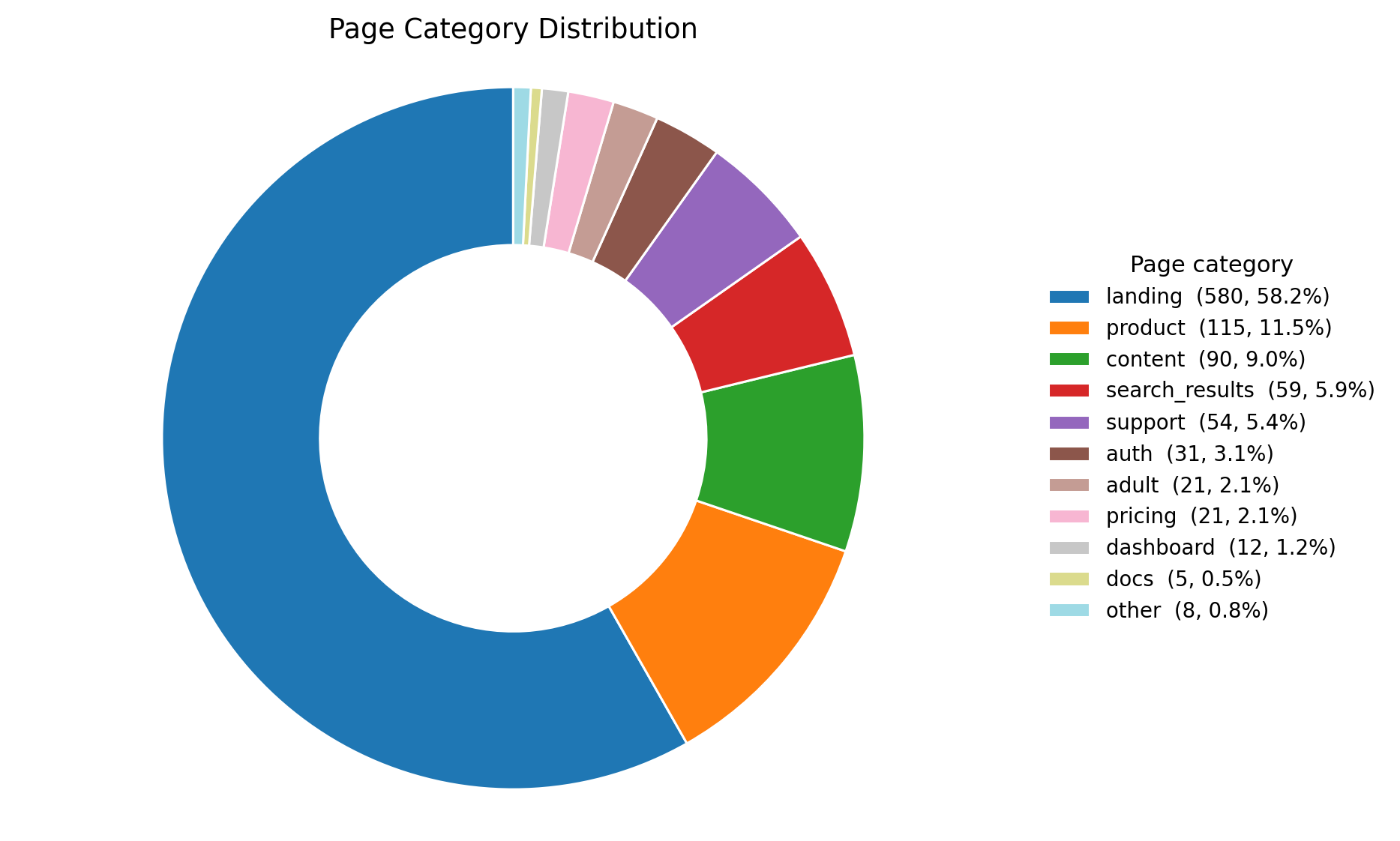}
    \caption{Page category distribution in \textsc{GuideWeb}.}
    \label{fig:page_category_dist_app}
\end{figure*}

\begin{figure*}[!ht]
    \centering
    \includegraphics[width=\textwidth]{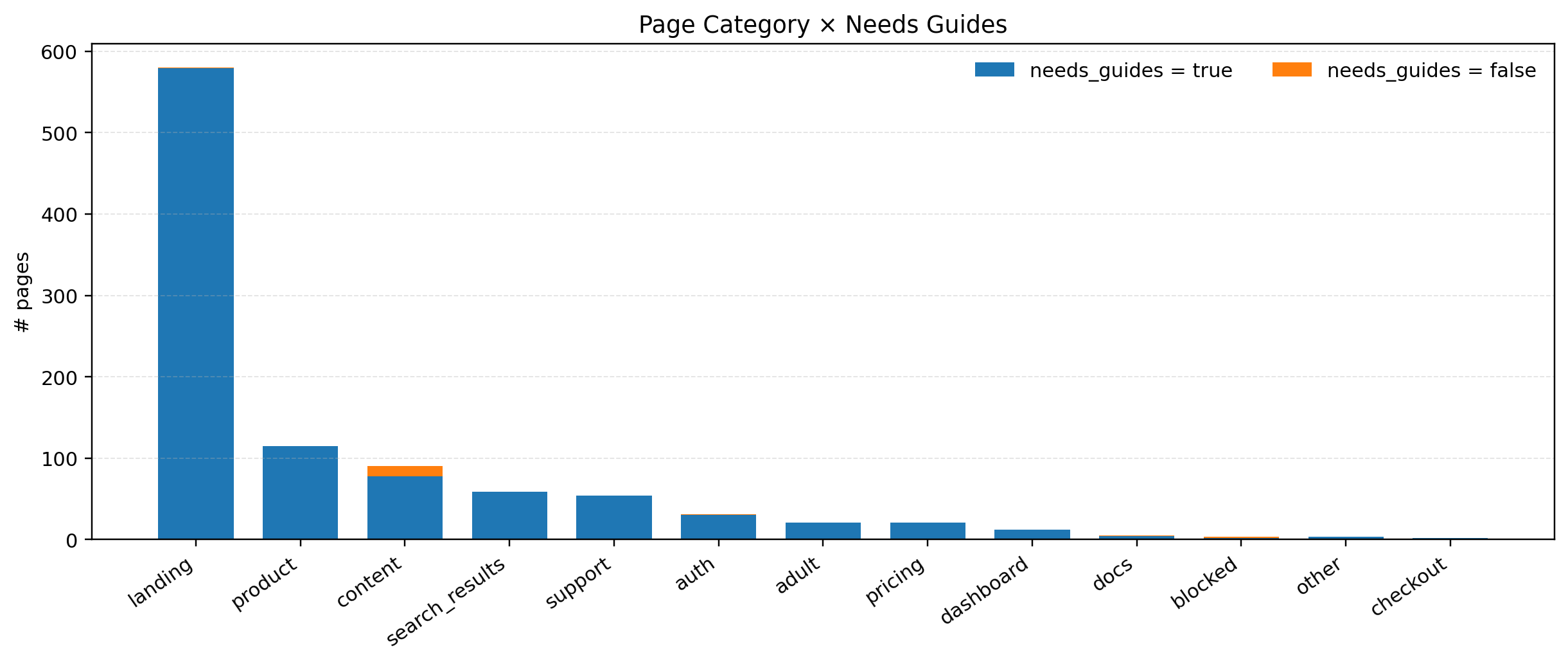}
    \caption{Page category by whether in-app guides are needed in \textsc{GuideWeb}.}
    \label{fig:page_category_needs_guides_app}
\end{figure*}

This appendix provides additional quantitative analyses of page categories in the \textsc{GuideWeb} benchmark.
Figure~\ref{fig:page_category_dist_app} presents the distribution of page categories, highlighting the dominance of landing pages alongside a diverse set of real-world entry points.
Figure~\ref{fig:page_category_needs_guides_app} further breaks down each category by whether pages require in-app guides, showing that only a small fraction of pages are labeled as not guide-worthy, and that such cases are concentrated in a few content-oriented categories.

\section{Metric Definitions}
\label{app:metrics}

\paragraph{Guide target selection.}
For a given webpage, let $G$ be the set of gold guide targets and $\hat{G}$ the predicted targets, where each target corresponds to a unique DOM element.
We define precision, recall, and F1 as follows:
\begin{equation}
\mathrm{Precision} \;=\; \frac{|G \cap \hat{G}|}{|\hat{G}|}.
\end{equation}
\begin{equation}
\mathrm{Recall} \;=\; \frac{|G \cap \hat{G}|}{|G|}.
\end{equation}
\begin{equation}
\mathrm{F1} \;=\; \frac{2\,\mathrm{Precision}\,\mathrm{Recall}}{\mathrm{Precision}+\mathrm{Recall}}.
\end{equation}

\paragraph{Field-level exact-match F1.}
For each structured field (e.g., \texttt{action\_type}, \texttt{tag}, \texttt{visible\_text}, \texttt{xpath}), we compute exact-match precision, recall, and F1 over field values aggregated across annotations.
Let $S$ denote the (multi)set of gold field values and $\hat{S}$ the (multi)set of predicted values.\footnote{Our implementation handles duplicates consistently between $S$ and $\hat{S}$.}
We define:
\begin{equation}
\mathrm{Precision} \;=\; \frac{|S \cap \hat{S}|}{|\hat{S}|}.
\end{equation}
\begin{equation}
\mathrm{Recall} \;=\; \frac{|S \cap \hat{S}|}{|S|}.
\end{equation}
\begin{equation}
\mathrm{F1} \;=\; \frac{2\,\mathrm{Precision}\,\mathrm{Recall}}{\mathrm{Precision}+\mathrm{Recall}}.
\end{equation}

\section{Implementation Details}
\label{appendix:implementation}

\paragraph{Baseline Inference Settings.}
All baseline models are evaluated in an inference-only setting without any task-specific fine-tuning.
We use the original model parameters and default decoding configurations provided by each model implementation.
To ensure fair comparison across models and to cover the majority of real-world webpages, we set the maximum input length to 130K tokens for all baselines.
This context window is sufficient to accommodate full HTML source code for most main pages in the \textsc{GuideWeb} benchmark after preprocessing, avoiding aggressive truncation that may remove critical interactive elements.
All baselines are prompted using a unified instruction template and generate guide target predictions and guide text in a single forward pass.

\paragraph{GuideWeb Agent Training.}
The \textsc{GuideWeb Agent} is trained using the \textsc{GuideWeb} training split with full-parameter fine-tuning.
Training is conducted on an NVIDIA GB10 GPU with 120\,GB of memory, which allows stable optimization with long-context inputs and structured output constraints.
We adopt a supervised fine-tuning (SFT) paradigm, where the model is trained to jointly predict guide targets and corresponding guide text conditioned on preprocessed webpage representations.

To reduce unnecessary computation caused by redundant or non-informative HTML content, we employ a \emph{shorter} mechanism that selectively retains interactive elements, visible text, and local contextual cues surrounding candidate targets.
This mechanism significantly shortens the effective input length while preserving task-relevant information, resulting in faster training and inference without degrading accuracy, as confirmed by our experimental results.

\paragraph{Training Hyperparameters.}
We fine-tune the \textsc{GuideWeb Agent} with full-parameter supervised learning using AdamW.
We train for \textbf{3 epochs} with a learning rate of \textbf{$1\times10^{-5}$}, warmup ratio \textbf{0.03}, and weight decay \textbf{0.0}.
The per-device batch size is \textbf{1} with gradient accumulation of \textbf{4} (effective batch size \textbf{4}).
We log every \textbf{10} steps and save checkpoints every \textbf{2000} steps.
For numerical stability and memory efficiency, we load the base model in \textbf{FP16} and train with \textbf{BF16} mixed precision, enabling \textbf{gradient checkpointing}.
We set the maximum prompt length to \textbf{3950} tokens and the maximum generation length to \textbf{2000} tokens, resulting in a maximum sequence length of \textbf{6014} tokens (including a small buffer).
All runs use a fixed random seed (\textbf{13}).

\paragraph{Input Construction and Shorter Configuration.}
We render each webpage using a fixed viewport of \textbf{1280$\times$720}.
For visible text, we keep up to \textbf{800} text blocks, truncating each block to at most \textbf{400} characters.
We additionally include a short global excerpt of the full-page text using an excerpt ratio of \textbf{0.02}, bounded to \textbf{[100, 200]} characters.
For structural cues, we keep headings (enabled) with at most \textbf{20} headings and at most \textbf{40} characters per heading.
For interactive candidates, we extract up to \textbf{2000} interactive elements, truncate each element's visible text to \textbf{120} characters, and enforce an overall interactive-text budget of \textbf{6500} characters.
To represent targets robustly, we provide XPaths using the \texttt{stable\_then\_abs} mode, limiting XPath textual fields to \textbf{80} characters and attribute fields to \textbf{200} characters.
Model outputs are required to end with a fixed end marker (\texttt{</JSON>}).

\paragraph{Filtering Long Samples.}
To avoid near-limit truncation and unstable supervision, we filter overly long samples during training.
Specifically, we drop samples whose estimated total tokens exceed \textbf{5200}, or whose prompt/output lengths are within \textbf{98\%} of the configured maximums.
Dropped instances are recorded in \texttt{dropped\_samples.jsonl} for traceability.

\end{CJK}
\end{document}